\documentclass{article}

\usepackage{arxiv}

\usepackage[utf8]{inputenc} 
\usepackage[T1]{fontenc}    
\usepackage{hyperref}       
\usepackage{url}            
\usepackage{booktabs}       
\usepackage{amsfonts}       
\usepackage{nicefrac}       
\usepackage{microtype}      
\usepackage{lipsum}
\usepackage{graphicx}
\usepackage{booktabs}
\usepackage{multirow}
\usepackage[normalem]{ulem}
\useunder{\uline}{\ul}{}

\usepackage{color}

\title{Fruit Quality and Defect Image Classification with Conditional GAN Data Augmentation}

\author{
  Jordan J. Bird\thanks{https://jordanjamesbird.com/} \\
  ARVIS Lab\\
  Aston University\\
  Birmingham, United Kingdom \\
  \texttt{birdj1@aston.ac.uk} \\
   \And
 Chloe M. Barnes\thanks{https://www.chloembarnes.com/} \\
  College of Engineering and Physical Sciences\\
  Aston University\\
  Birmingham, United Kingdom \\
  \texttt{barnecm1@aston.ac.uk} \\
     \And
 Luis J. Manso\thanks{https://ljmanso.com/} \\
  ARVIS Lab\\
  Aston University\\
  Birmingham, United Kingdom \\
  \texttt{l.manso@aston.ac.uk} \\
       \And
  Anik\'o Ek\'art \\
  College of Engineering and Physical Sciences\\
  Aston University\\
  Birmingham, United Kingdom \\
  \texttt{a.ekart@aston.ac.uk} \\
     \And
  Diego R. Faria\thanks{https://cs.aston.ac.uk/$\sim$fariad/} \\
  ARVIS Lab\\
  Aston University\\
  Birmingham, United Kingdom \\
  \texttt{d.faria@aston.ac.uk} \\
}

\begin{document}
\maketitle

\begin{abstract}
Contemporary Artificial Intelligence technologies allow for the employment of Computer Vision to discern good crops from bad, providing a step in the pipeline of selecting healthy fruit from undesirable fruit, such as those which are mouldy or gangrenous. State-of-the-art works in the field report high accuracy results on small datasets (<1000 images), which are not representative of the population regarding real-world usage. The goals of this study are to further enable real-world usage by improving generalisation with data augmentation as well as to reduce overfitting and energy usage through model pruning. In this work, we suggest a machine learning pipeline that combines the ideas of fine-tuning, transfer learning, and generative model-based training data augmentation towards improving fruit quality image classification. A linear network topology search is performed to tune a VGG16 lemon quality classification model using a publicly-available dataset of 2690 images. We find that appending a 4096 neuron fully connected layer to the convolutional layers leads to an image classification accuracy of 83.77\%. We then train a Conditional Generative Adversarial Network on the training data for 2000 epochs, and it learns to generate relatively realistic images. Grad-CAM analysis of the model trained on real photographs shows that the synthetic images can exhibit classifiable characteristics such as shape, mould, and gangrene. A higher image classification accuracy of 88.75\% is then attained by augmenting the training with synthetic images, arguing that Conditional Generative Adversarial Networks have the ability to produce new data to alleviate issues of data scarcity. Finally, model pruning is performed via polynomial decay, where we find that the Conditional GAN-augmented classification network can retain 81.16\% classification accuracy when compressed to 50\% of its original size. 
\end{abstract}


\section{Introduction}
Recognition of fruit quality is important in smart agriculture to increase production efficiency. Contemporary Artificial Intelligence technologies allow for the employment of Computer Vision to discern good crops from bad, providing a step in the pipeline of selecting healthy fruit from undesirable fruit, such as those which are mouldy or gangrenous. 

Even during modern times, collecting a dataset that generally represents a species of fruit poses difficulties. For example, the widely used Fruits 360 dataset~\cite{murecsan2018fruit} contains 2134 images of apples belonging to one of thirteen cultivars. According to the United Nations, 87.2 million tonnes of apples were farmed globally in 2019 alone~\cite{faostat_2021} belonging to over 7500 cultivars~\cite{elzebroek2008guide}. In terms of smart agriculture, this highlights the issue of data scarcity. On the problem of crop quality recognition, much data is required to generalise to a population and thus become apt for real-world use. Considering the yield of fruit globally compared to practical data collection, bridging this gap manually i.e., \textit{collecting more data}, is simply an impossible task. 

In this work, we focus on exploring a solution to this problem for lemon harvesting. Recording data for both lemons and limes, the United Nations noted that 20 million tonnes were harvested in 2019~\cite{faostat_2021}. According to the CIA World Factbook~\cite{cia_exports}, lemon and lime fruit exports comprised over \$3.3 billion USD of international exports in 2019. The largest exporters were Spain (\$828 million USD), Mexico (\$523 million USD), and the Netherlands (\$339 million USD). Although the top 15 countries exported 93.5\% of all lemons and limes in 2019, many countries are expanding their efforts; for example, Belize increased lemon and lime exports by 11,100\% from 2018 to 2019, followed by Timor-Leste and Georgia which increased exports by 4,200\% and 2,115\%, respectively. With this information in mind, the ability to autonomously select and reject lemon fruit based on health, e.g., sorting out those which have developed mould or gangrene, would allow for further increases in an already growing market by increasing production efficiency.

In this article, we explore a Conditional Generative Adversarial Network-based solution to data scarcity through training data augmentation. The main scientific contributions we present in chronological order are as follows:
\begin{itemize}
    \item Exploration of Convolutional Neural Network (CNN) topologies for fruit quality recognition.
    \item Implementation of a Conditional Generative Adversarial Network (Conditional GAN) for the generation of synthetic healthy and unhealthy fruit images. The trained synthetic data generation model is made available for future work\footnote{Image generation model weights and code are available at:\\ \url{https://github.com/jordan-bird/synthetic-fruit-image-generator}}.
    \item Augmentation of the original dataset with synthetic images to improve the CNN performance.
    \item Exploration of features within synthetic images shows that the Conditional GAN learns to generate healthy synthetic fruit images with no defects, as well as unhealthy synthetic fruit images with defects such as mould and gangrene. 
    \item Model pruning shows that a Conditional GAN-augmented classification network can retain 81.16\% classification accuracy when compressed to 50\% of its original size. 
\end{itemize}

\section{Background and Related Work}
\subsection{Fruit Quality Recognition}
Fruit Quality Recognition is a technique where a fruit can be scored or classified autonomously by an algorithm given input features such as photographs. As previously mentioned, the ability to perform this task autonomously (and non-invasively~\cite{vetrekar2015non}) allows for an increase of production efficiency since sorting can be performed by machines endowed with such an algorithm. 

Reduction in cost of such a system is of particular interest in the field given that several Lower Economically Developed Countries (LEDCs) are expanding the production and export of lemon fruit~\cite{cia_exports}. Solutions such as electronic noses can cost up to 100,000\$ USD~\cite{chang2008electronic}, whereas a camera and computer are a fraction of the cost, arguing that image recognition is a more viable option when cost is an issue. It is worth noting though, that low-cost electronic noses are currently an expanding line of research within the Sensors and Internet of Things (IoT) fields~\cite{garcia2019low}. Electronic noses are indeed strongly performing solutions to fruit quality recognition as shown by \cite{brezmes2001correlation}, \cite{di2001evaluation}, and \cite{brezmes2005evaluation}.

Deep learning approaches to fruit classification are abundant, as shown by multiple literature reviews on the subject~\cite{zawbaa2014automatic,dubey2015application,naik2017machine,hameed2018comprehensive,naranjo2020review}. In comparison, the application of deep learning to the more fine-grained problem of quality classification are relatively rare. Bhargava and Bansal show that the field of automated fruit quality recognition is rapidly growing, in part due to the availability of newer technologies~\cite{bhargava2018fruits}. 

Given that one can discern quality of a fruit based on observation, visual features are often noted as of importance when it comes to fruit quality classification. Yamamoto et al.~\cite{yamamoto2015strawberry} suggested that Linear Discriminant Analysis (LDA) of colour, shape, and size enabled the formation of a distance matrix which could be used to classify both the cultivar and quality of strawberries. Usage of a single LDA led to a classification accuracy of 42\% whereas combining the three analyses caused accuracy to rise to 68\%. Capizzi et al. followed a similar technique through texture and gray-features with a Radial Basis Probabilistic Neural Network, which scored around 97.25\% on a limited set of images of orange fruits~\cite{capizzi2015automatic}. Azizah et al.~\cite{azizah2017deep} provided a solution to defect classification in the Mangosteen fruit, attaining a mean 97.5\% 4-fold classification accuracy, albeit with a limited dataset. This pattern of data scarcity continues as would be expected in the field; in 2020, Fan, et al.~\cite{fan2020line} found that CNNs could classify apple defects with around 96.5\% accuracy after processing 300 fruit images (150 per class). This study also notes the efficiency of deep learning algorithms post-training, the algorithm was capable of processing 5 fruit images per second (0.2 seconds each).

In this work, we take inspiration from Osako et al.'s approaches to cultivar discrimination of lychee fruit~\cite{osako2020cultivar}. The study showed success of fruit image classification (albeit for a different task of cultivar recognition) when fine-tune transfer learning with the VGG16 CNN~\cite{simonyan2014very}, and predictions were analysed super-imposed upon the images via Grad-CAM~\cite{selvaraju2017grad} in order to explain useful features for discrimination. We follow a similar approach in this work (applied to a new problem) in terms of fine-tuning of VGG16 and analysis with Grad-CAM, and go a step further in improving classification through data augmentation with a Conditional Generative Adversarial Network to self-regularise the network by creating new, synthetic fruit images.

\subsection{Data Scarcity and Augmentation}
Regarding the global fruit yield statistics versus dataset size examples within the introduction, data scarcity in machine learning is the reliance of models on exhaustive labelling, providing a limitation to their real-world use~\cite{zhang2020towards}. Given that the use of a model is to aim towards generalisation of a population, a lack of data can lead to a situation wherein training accuracy scores are high and yet deployment to industry would lead to failure. As noted in the introduction, gathering enough data of cultivated fruit is impractical. Without enough data to properly represent the population, models will be prone to overfitting. Given this, other methods are required to prevent overfitting and encourage generalisation towards the real-world use of a machine learning model outside the realm of simply collecting more data. 

Data augmentation is the process of creating new training data by either slightly modifying the data at hand or generating new, synthetic data~\cite{shorten2019survey}. An augmented dataset thus provides more training examples for a given task. 

Image recognition tasks for Convolutional Neural Network image classification are affected by data scarcity due to their data requirements~\cite{bloice2017augmentor,andriyanov2020using}, where many generative models have been recommended to alleviate such issues~\cite{nalepa2019data,tran2021data}. Generative models have also been noted to positively impact biological signal classification~\cite{anicet2020parkinson,bird2021synthetic}, semantic Image-to-Image Translation~\cite{arantes2020csc}, speech processing~\cite{qian2019data,bird2020overcoming}, and Human Activity Recognition~\cite{alnujaim2019generative,erol2019gan} among many others. In this work, we use a Conditional GAN for data augmentation, which are described in the following section. This is based on the literature wherein GANs and Conditional GANs have been noted to perform particularly well in image classification~\cite{frid2018gan,han2019combining,lee2019conditional,loey2020deep}. We note specific inspiration from Fu et al.~\cite{fu2020multi}, where Conditional GANs have been noted to perform well on fine-grained images such as classification of bird and dog breeds (rather than classification of a whole species). This bares similarity to our problem, where finer details on generally similar images dictate which class they belong to.

\subsection{GAN and Conditional GAN}
The Generative Adversarial Network (GAN) was first introduced in 2014~\cite{ganoriginal}. The idea behind the GAN is to have two neural networks compete in a zero-sum game ergo \textit{adversarial}, i.e., the loss of one network is directly beneficial to the other and vice versa. To give an example of image generation, as this work performs, there are two networks; a generator network which creates images, and a discriminator network which classifies the inputs as either real or fake. As with most deep learning approaches, the gradients of each network are updated after each training batch with a stochastic gradient algorithm. The output of the generator network feeds directly into the discriminator network and thus training of the two networks is automated via their competition. In terms of categorical cross-entropy, a score can be calculated as follows:
\begin{equation}
\label{eq:gan}
    E_{x}[log(D(x))] + E_{z}[log(1-D(G(z)))],
\end{equation}
where the first part of the equation ($E_{x}[log(D(x))]$) is the recognition of real images and the second part ($E_{z}[log(1-D(G(z)))]$) is the recognition of fake images. $E_{x}$ and $E_{z}$ are the expected values over all real and fake data, respectively; e.g., $x$ is a real input from the dataset and $z$ may begin as a random noise input to a generator. The function $D(x)$ is the probability that a given data is real and is thus therefore being reversed to discern fake images. Note that $D(x)$ is replaced by $G(z)$ in the second part of the equation, this is due to input to the discriminator being Generator $G$'s output when presented with random input vector $z$. This is known as a \textit{minimax loss}, since the generator's aim is to maximise Equation \ref{eq:gan} while the discriminator aims to minimise it.

A Conditional Generative Adversarial Network (CGAN or Conditional GAN)~\cite{mirza2014conditional} is an extension of the above technology, but with a given class label. That is, the generator now aims to learn to generate images belonging to one of $n$ classes, in this work this is a binary label of ``healthy" and ``unhealthy". Equation \ref{eq:gan} can be extended as follows:
\begin{equation}
\label{eq:cgan}
    E_{x}[log(D(x|y))] + E_{z}[log(1-D(G(z|y)))],
\end{equation}
where data objects $x$ and $D(z)$ are given class label $y$. Therefore, $D(x|y)$ is the discriminator's probability that $x$ is real given class label $y$, and $G(z|y)$ is the output of the generator with random vector $z$ given class label $y$. 
\begin{figure}
    \centering
    \includegraphics{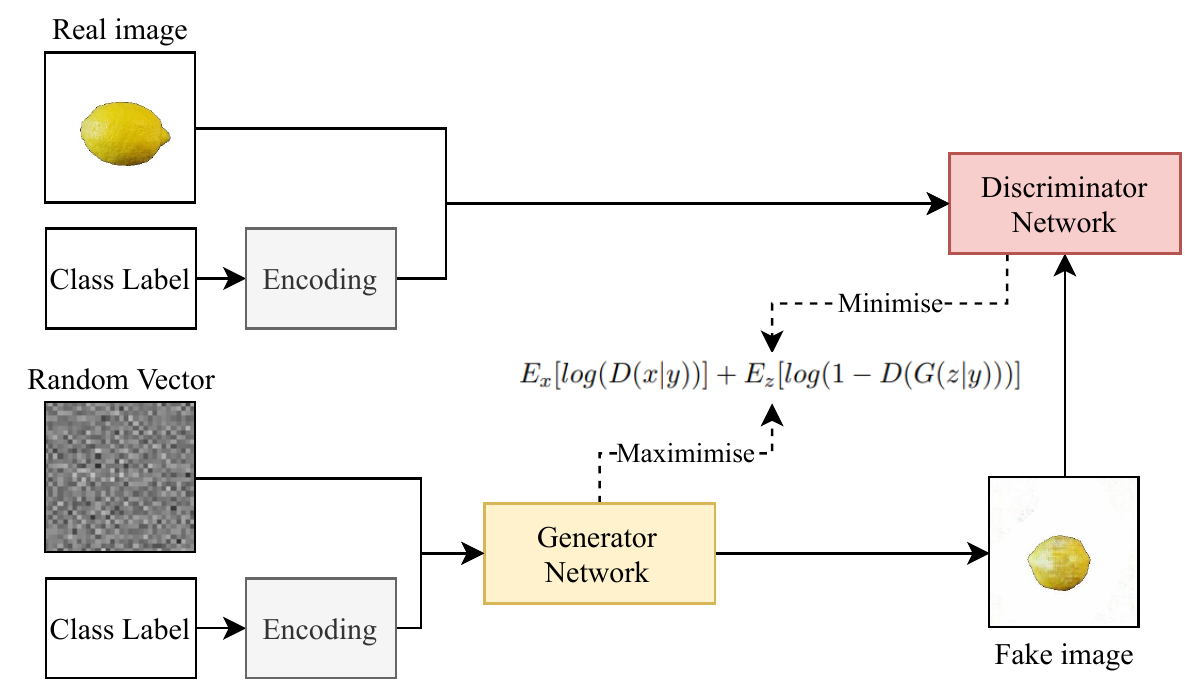}
    \caption{Generalisation of a Conditional GAN topology.}
    \label{fig:cgantopo}
\end{figure}
This minute difference in topology from a GAN, as can be observed in Figure \ref{fig:cgantopo}, allows for the generation of objects belonging to multiple classes. If the dataset in this work was presented to a vanilla GAN, the network would learn to generate fake fruit images by learning from real fruit, thus two networks would then be needed for the generation of either class. Said networks would have to train independently of one another. By using a Conditional GAN, we can specify to the network whether we want it to generate healthy or unhealthy fruit by learning not only to generate them in the general sense, but also learning from the significance of a class label. \\

\section{Method}
\begin{figure}
    \centering
    \includegraphics{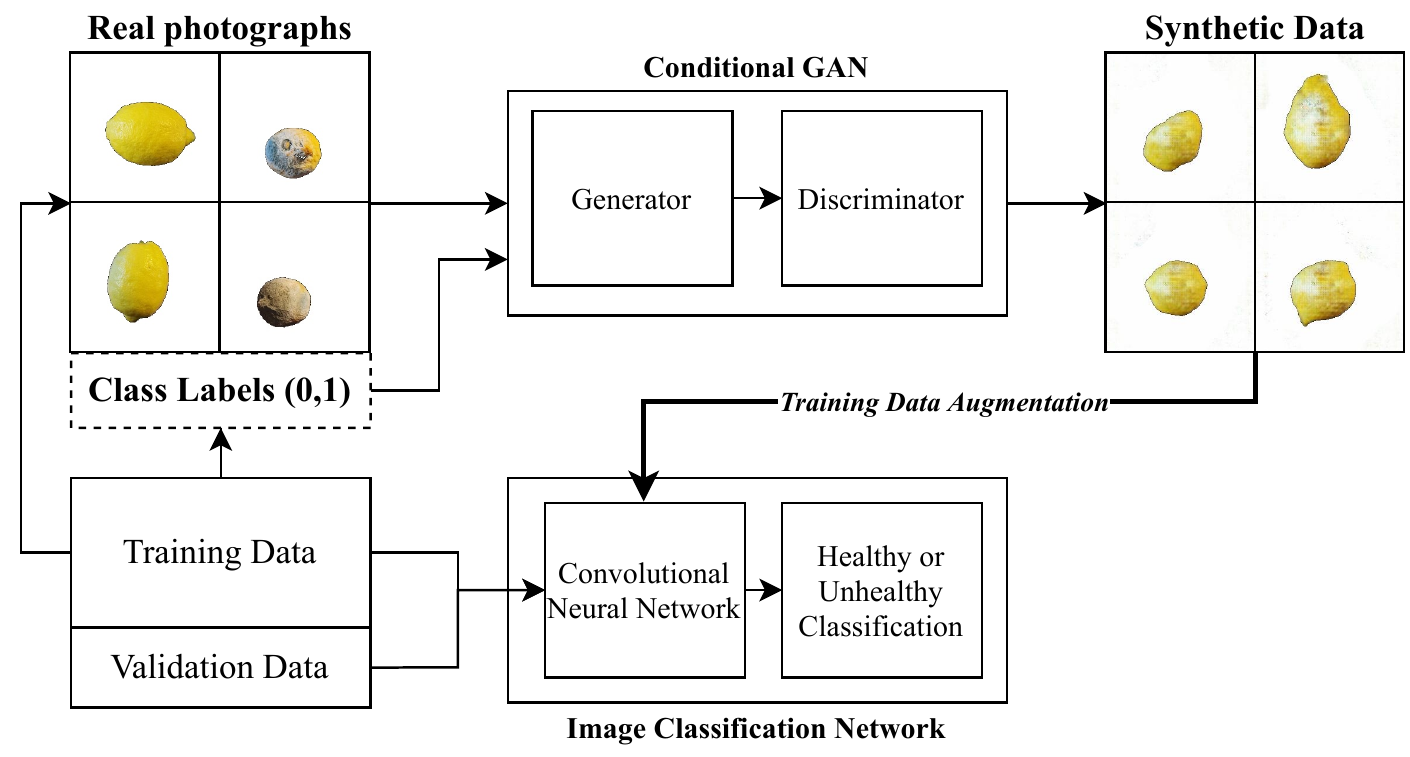}
    \caption{A general overview of our proposed approach for data augmentation towards fruit quality image classification.}
    \label{fig:general_diagram}
\end{figure}
A general overview of the proposed approach can be observed in Figure \ref{fig:general_diagram}, where the training data is augmented with a Conditional GAN approach. In this section, we describe the method for each step of the experiments performed. 

\subsection{Data Collection and Preprocessing}
Initially, an open source dataset of lemon images were acquired from SoftwareMill~\cite{softwaremill_2020}\footnote{Note: None of the authors of this work are affiliated with SoftwareMill}. The dataset contains 2690 images of lemons at a resolution of 1056$\times$1056 pixels and are annotated in COCO format. Given that each COCO annotation describes one class, i.e., a fruit that exhibits both mould and gangrene will have two individual entries, we sort through the dataset to apply a single binary class label of ``healthy" or ``unhealthy" to each of the fruit images. 

Given the computational complexity of the algorithms involved, the images are then resized to 256$\times$256 pixels; this resolution still allows for the visualisation of undesirable features while reducing the total number of RGB pixel values from 3,345,408 (1056$\times$1056$\times$3) to 196,608 (256$\times$256$\times$3). This reduction to 5\% of the original model inputs reduces the amount of memory required for training of all models, since the use of full resolution images is not feasible for consumer-grade hardware. In terms of deployment and real-world usage, robots themselves will have energy restrictions due to the processing cost and profit tradeoff regarding automation of fruit sorting. Thus, this reduction in image size increases the practicality of the approach.

To better discern noise throughout the generative learning process, the black background is replaced with white. Though it would have no effect on the training process of the model, the background is replaced so visual glitches can be better discerned through manual observation throughout training. For example, later in Figure \ref{fig:epoch_comparisons}, several small glitches occur in the eighth generation of outputs that would have been more difficult to observe in the presence of a dark background.

\begin{figure}
    \centering
    \includegraphics[scale=0.85]{"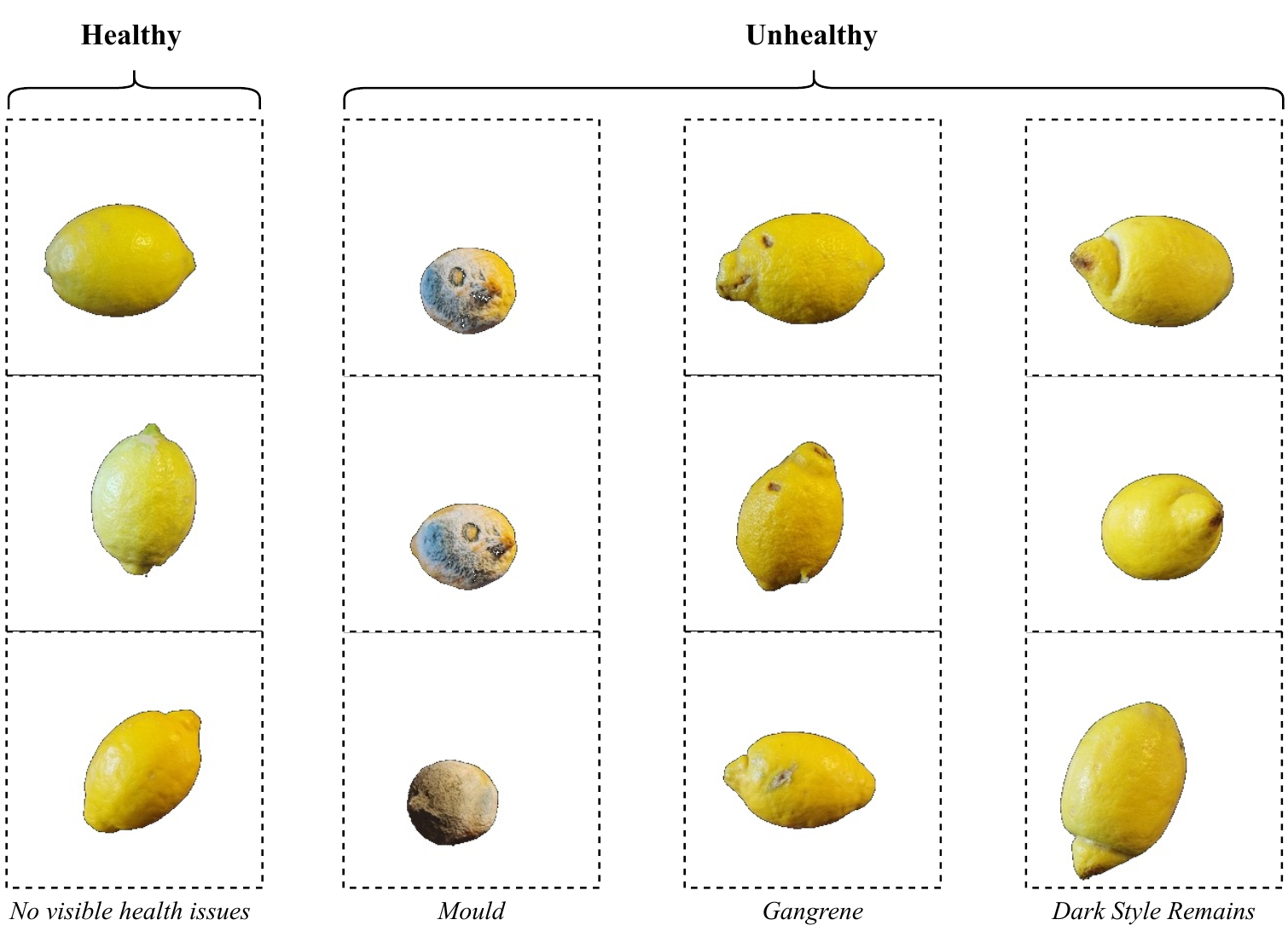"}
    \caption{Visualisation of healthy and unhealthy lemons within the dataset. Mouldy, gangrenous, and those with a dark style remaining are all considered unhealthy.}
    \label{fig:comparison_photos}
\end{figure}
Examples of some images in the preprocessed dataset can be found in Figure \ref{fig:comparison_photos}. Healthy lemons are gathered for the healthy class, whereas mouldy, gangrenous, and those with a dark style remaining are gathered to form the unhealthy class. 

\subsection{Data Augmentation via Conditional GAN}
For data augmentation, a Conditional GAN is utilised. The model is selected since it supports the concatenation of the generator and discriminator networks with a second input of the class label. That is, the context of a healthy or unhealthy fruit is specified, and so the model will learn to generate images as belonging to either one of the two classes.

The initial input to the generator is a vector representing a three-channel 8$\times$8 pixel image (8$\times$8$\times$3). 
By using Convolutional Transpose layers, this is eventually upscaled to a 256$\times$256 RGB image. The discriminator network downsamples twice with convolutional layers of 128 neurons, a kernel size of $(3,3)$ and a stride of $(2,2)$. Each layer utilises LeakyReLU activation~\cite{maas2013rectifier} whereas the output is set as a hyperbolic tangent for scaling, and the ADAM optimiser~\cite{kingma2014adam} is used to train. Latent space for class label interpretation is of size 100. Hyperparameter selections are based on the findings of the studies in \cite{radford2015unsupervised}. 

The Conditional GAN was first initially trained for 500 epochs, manual exploration of the produced synthetic data showed promise, but several severe visual glitches still occurred. Due to this, the training was extended and performed for 2000 epochs in total with a batch size of 64. It was also observed that batch sizes below 64 caused the generator to cease training after around 10 epochs and failed to learn any further. 

\subsection{Classification, Model Analysis and Pruning}
\begin{figure}
    \centering
    \includegraphics{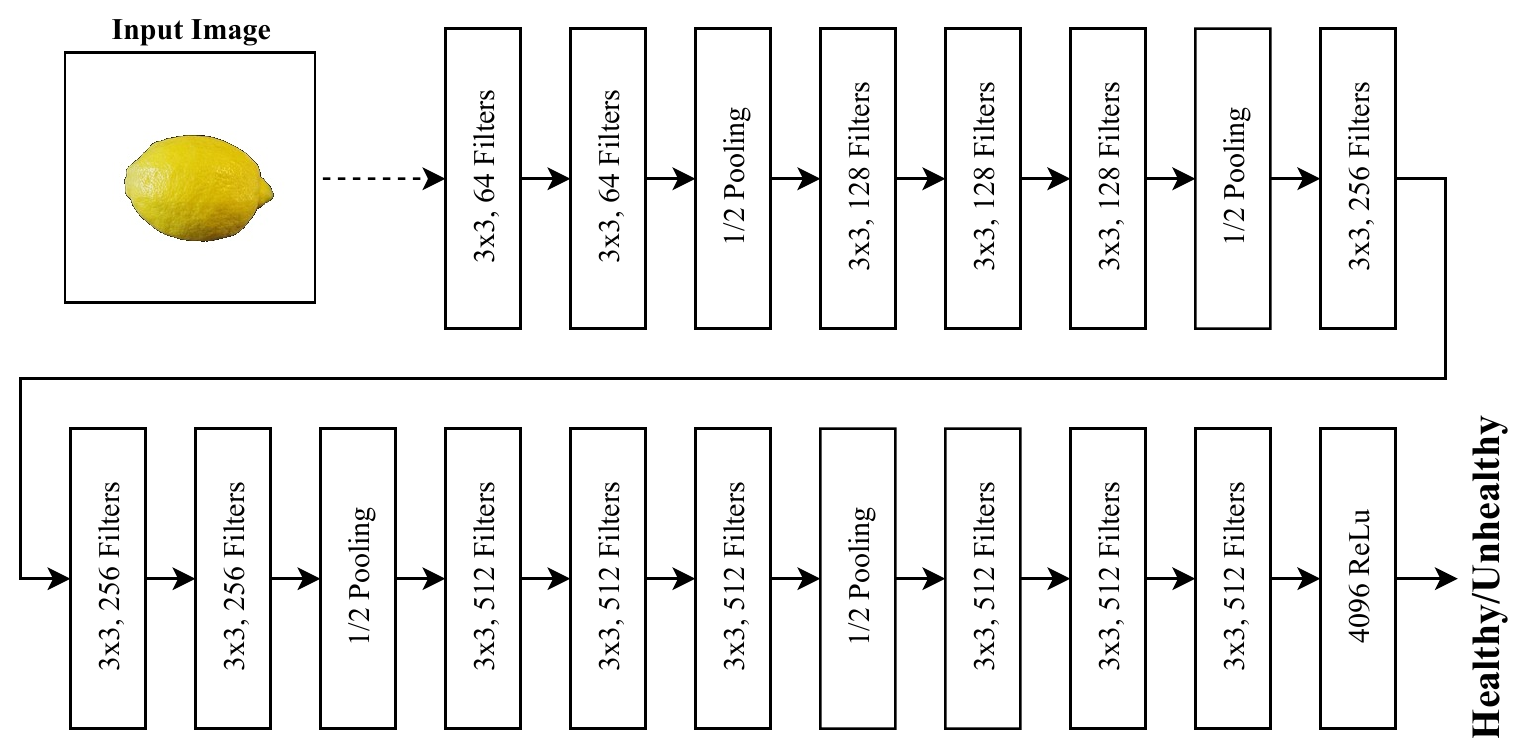}
    \caption{Overview of the VGG16 topology and custom interpretation and output layer which is used for binary image classification.}
    \label{fig:vgg16}
\end{figure}
The image classification network itself utilises the concept of fine-tune transfer learning from a large ImageNet-trained model, VGG16~\cite{simonyan2014very}. A diagram of the Convolutional Neural Network topology we use for the classification of fruit quality images can be observed in Figure \ref{fig:vgg16}. The final three ReLu layers and SoftMax predictions have been replaced by a single interpretation layer and a single output neuron with a sigmoid activation function for the optimisation of binary cross-entropy.

The number of neurons within the interpretation layer is optimised through a linear search of (8, 16, 32, 64, ..., 8192) neurons. The network is given a maximum of 100 epochs to train, but training is stopped early if no further learning occurs within a time of 10 epochs. 

Explainiability in AI is important, especially when such algorithms are considered for real world usage. Algorithms tend to operate in a black-box like nature, for example, the algorithm in this work would take as input an image of a lemon fruit and produce a class label output, corresponding to whether the fruit is healthy or not. With this in mind, regardless of the training accuracy scores attained, further analysis is needed to explain why decisions are made and predictions are given. We analyse several synthetic images through Gradient-weighted Class Activation Mapping (Grad-CAM)~\cite{selvaraju2017grad,huff2021interpretation}. Class activation maps are produced by the convolutional neural network trained only on real images when given synthetic data as input. This allows us to confirm that undesirable characteristics are indeed both generated and classified as being important; since a GAN generator's goal is simply to learn to outperform the discriminator, and this may be done through other means i.e. finding methods to \textit{trick} the classifier. 

Considering that processing is performed on large quantities of fruit and that there may be time and energy restrictions, model pruning is performed on the network to explore the possibility of smaller model sizes, more apt for real-world usage~\cite{molchanov2019importance,fountsop2020deep}, through polynomial decay. For each model, 9 pruning experiments were performed with weight sparsity ranging from 0.9 (10\% of original size) to 0.1 (90\% of original size). Pruning was performed on the whole model for 20 epochs, from a sparsity of 0 (full size) to the given value for the individual experiment (0, 0.1, 0.2, ..., 0.9).

\subsection{Experimental Software and Hardware}
The models in this work were implemented in the Keras library with a TensorFlow backend. Models were trained on an RTX 2080Ti GPU (4352 CUDA cores).

\section{Results}
\subsection{Non-Augmentation Results}
\begin{table}[]
\centering
\caption{A linear search for the best number of CNN interpretation neurons for the classification of the real dataset, repeated three times with different random seeds.}
\label{tab:cnn-neurons}
\begin{tabular}{@{}lllll@{}}
\toprule
\multirow{2}{*}{\textbf{\begin{tabular}[c]{@{}l@{}}CNN Output\\ Interpretation Neurons\end{tabular}}} & \multicolumn{4}{l}{\textbf{Validation Accuracy (\%)}}                                                         \\ \cmidrule(l){2-5} 
                                                                                                      & \textit{\textbf{Seed = 1}} & \textit{\textbf{Seed = 2}} & \textit{\textbf{Seed = 3}} & \textit{\textbf{Mean}} \\ \cmidrule(r){1-1}
\textit{\textbf{8}}                                                                                   & 60.1                       & 60.97                      & 57.13                      & 59.40                  \\
\textit{\textbf{16}}                                                                                  & 60.1                       & 78.31                      & 71.5                       & 69.97                  \\
\textit{\textbf{32}}                                                                                  & 81.41                      & 60.97                      & 57.13                      & 66.50                  \\
\textit{\textbf{64}}                                                                                  & 75.71                      & 78.93                      & 57.13                      & 70.59                  \\
\textit{\textbf{128}}                                                                                 & 81.04                      & 79.8                       & 76.33                      & 79.06                  \\
\textit{\textbf{256}}                                                                                 & 73.23                      & 75.84                      & 65.55                      & 71.54                  \\
\textit{\textbf{512}}                                                                                 & 81.91                      & 80.79                      & 79.06                      & 80.59                  \\
\textit{\textbf{1024}}                                                                                & 73.85                      & 83.02                      & 80.17                      & 79.01                  \\
\textit{\textbf{2048}}                                                                                & 83.4                       & 81.91                      & 79.68                      & 81.66                  \\
\textit{\textbf{4096}}                                                                                & 82.28                      & \textbf{83.77}             & 79.93                      & {\ul \textbf{81.99}}   \\ 

\textit{\textbf{8192}}                                                                                & 81.54                     & 81.78             & 81.41                      & 81.58   \\

\bottomrule
\end{tabular}
\end{table}
Table \ref{tab:cnn-neurons} shows the exploration of interpretation neurons following the VGG16 ImageNet Convolutional Neural Network. Following the three tests, the lowest observed accuracy was that from 8 interpretation neurons with a mean classification accuracy of 59.4\%. The best model was the network with 4096 interpretation neurons, which scored 81.99\% classification accuracy, and the best individual run was the second run of the 4096-neuron network which scored 83.77\% classification accuracy.

\subsection{Conditional GAN Training}
\begin{figure}
    \centering
    \includegraphics[scale=0.65]{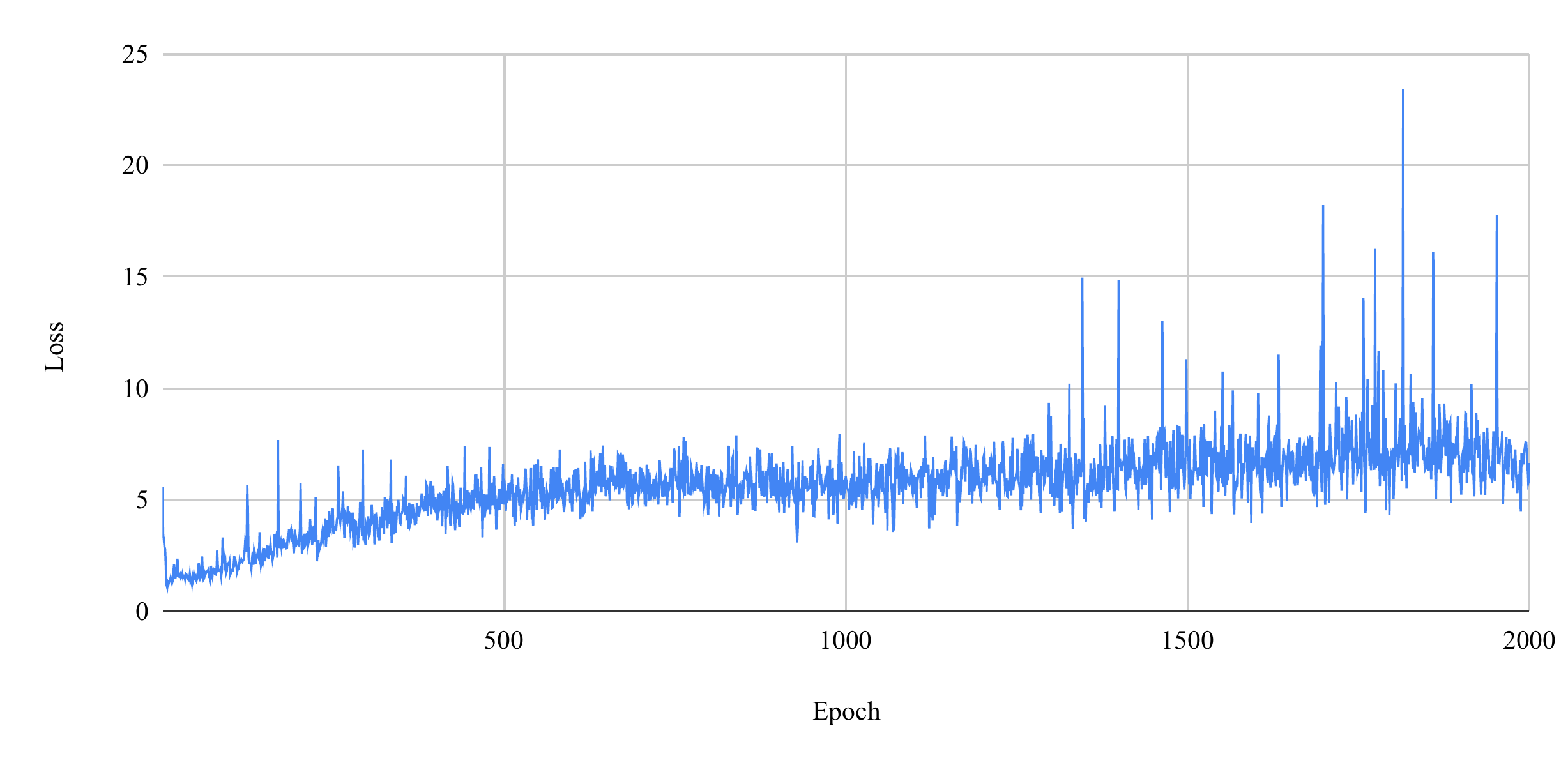}
    \caption{Observed loss (errors) for the generator network of the Conditional GAN during training.}
    \label{fig:generator_loss}
\end{figure}
\begin{figure}
    \centering
    \includegraphics[scale=0.65]{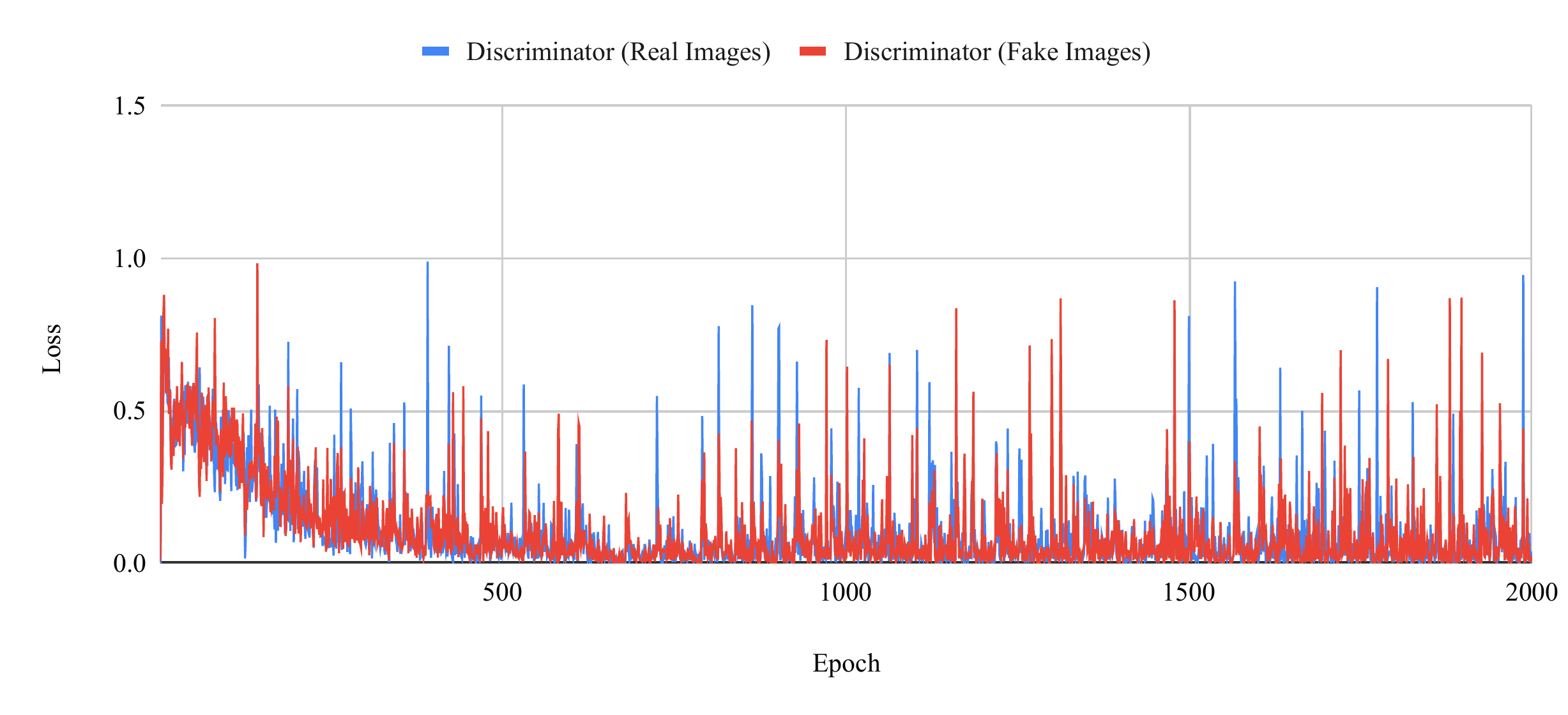}
    \caption{Observed loss (errors) for the discriminator network of the Conditional GAN during training. Loss is calculated for recognition of real and fake images separately.}
    \label{fig:discriminator_loss}
\end{figure}

Separated into two graphs for readability purposes, Figures \ref{fig:generator_loss} and \ref{fig:discriminator_loss} show the observed losses for the generator and discriminator networks of the Conditional GAN, respectively. Albeit with several anomalous spikes, it can be observed that the generator starts at a high loss of 5.5 which drops throughout the first few epochs. The discriminator, as can be expected, starts low given the quality of output by the generator. The generator can be seen to rise steadily for the first 500 epochs before becoming relatively stabilised. The two discriminator network losses were lower throughout these first 500 epochs before showing an oscillatory nature during the remainder of the learning process. 

\begin{figure}
    \centering
    \includegraphics[scale=0.9]{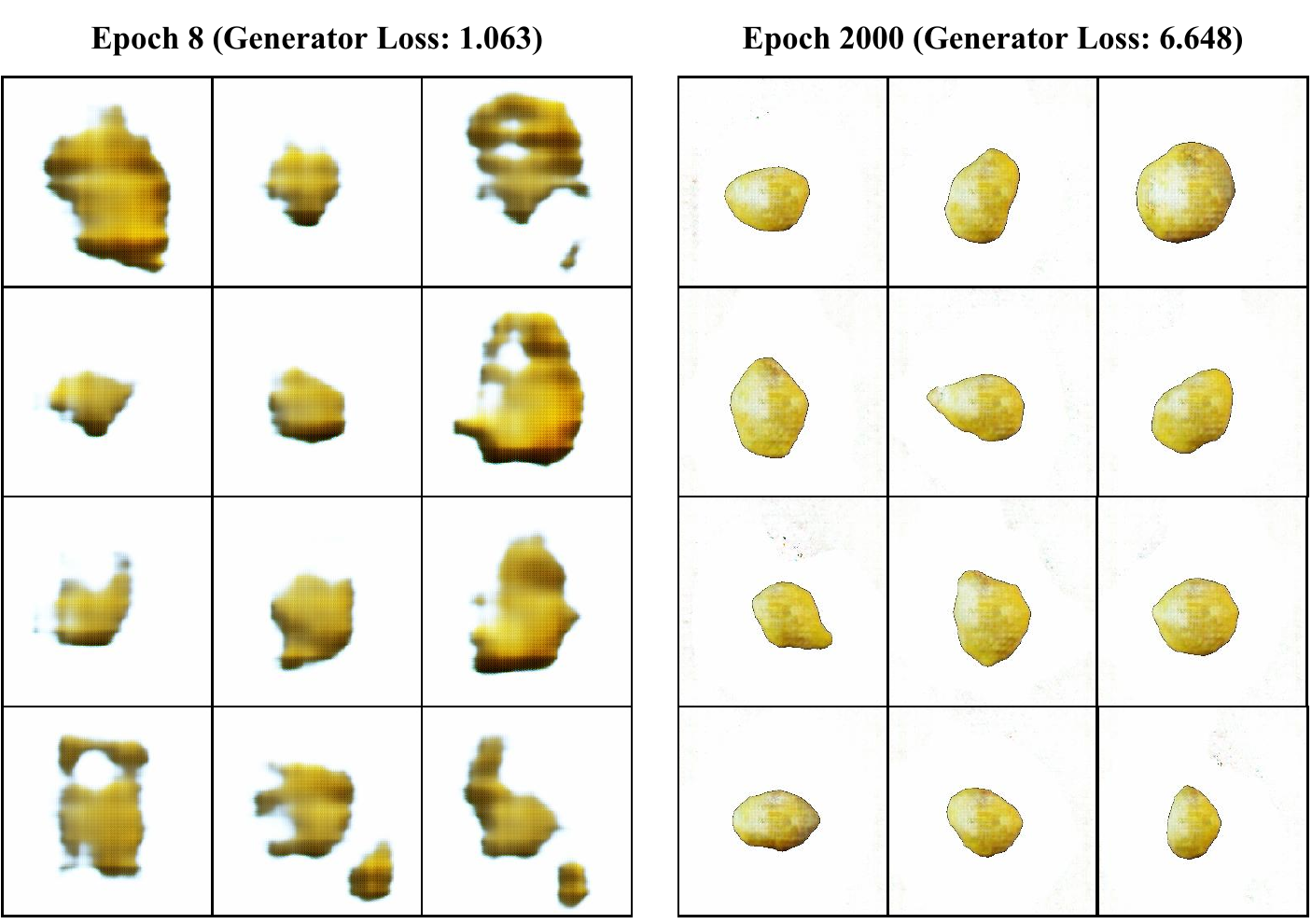}
    \caption{Examples of generator outputs at the lowest observed loss compared to the final epoch.}
    \label{fig:epoch_comparisons}
\end{figure}

At the end of the final epoch, the discriminator losses were 0.013 for the real images and 0.005 for the fake images produced by the generator. The generator loss was 6.648. Indeed, this value of 6.648 is by far not the lowest observed, but it is important to consider the nature of GANs; the losses of the two networks are relative to one another, i.e. it is an adversarial score. To provide an example of this, Figure \ref{fig:epoch_comparisons} shows a comparison of the images produced by the generator with the lowest loss (epoch 8 at 1.063) and then the generator at the final epoch. Evidently, the images produced by the final generator are of much higher quality than those output by the generator when it experiences the lowest observed loss. For this reason, the final generator is selected as the synthetic data producing model in this work, although it is suggested in the future to explore the quality at multiple stages.
\begin{figure}
    \centering
    \includegraphics[scale=0.8]{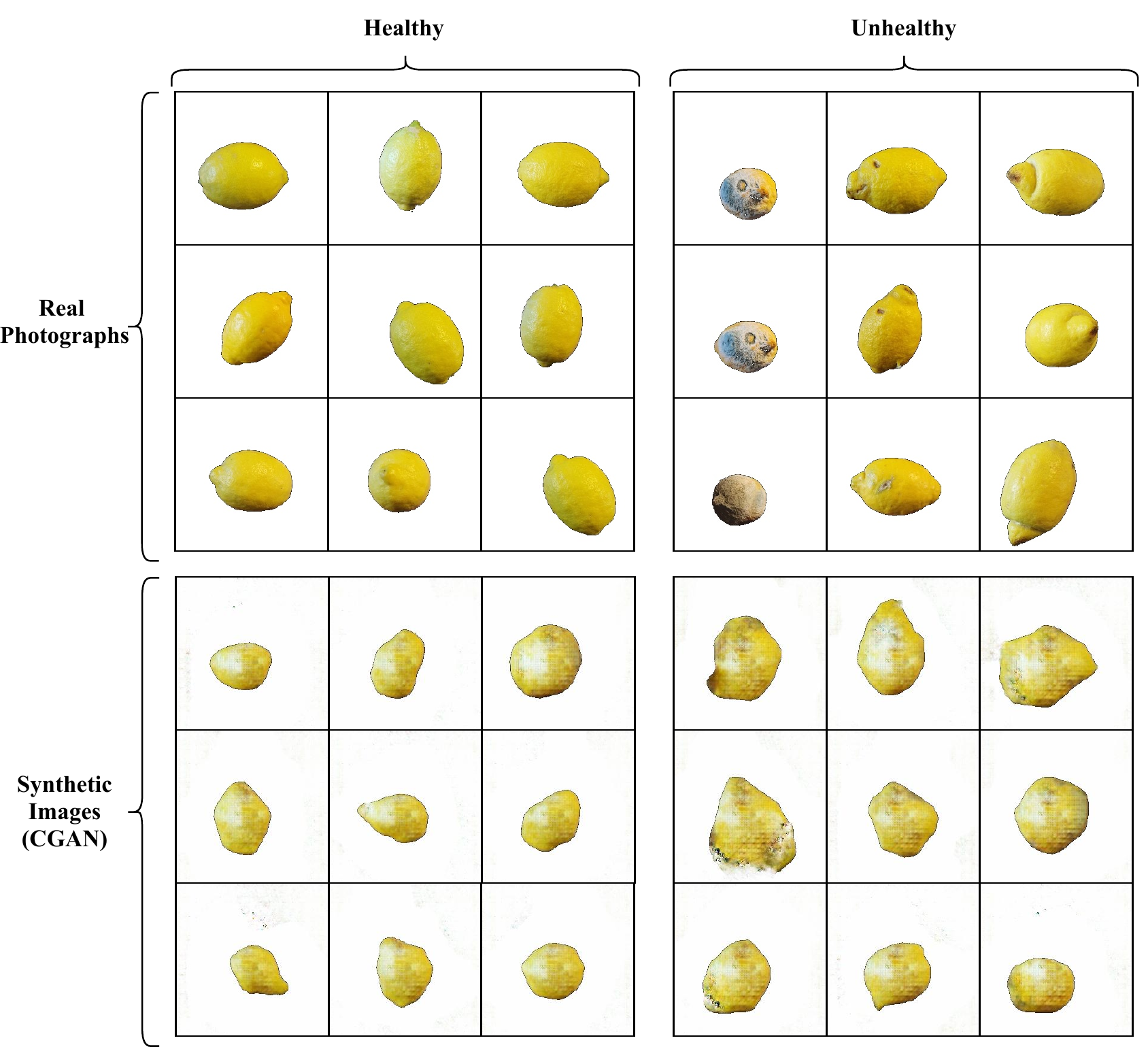}
    \caption{Examples of both real photographs and Conditional GAN outputs (Epoch 2000) for the two classes of healthy and unhealthy lemons.}
    \label{fig:real_gan_examples}
\end{figure}
Figure \ref{fig:real_gan_examples} shows 18 real photographs for healthy and unhealthy classed lemons, as well as 18 examples of Conditional GAN generator output for healthy and unhealthy classed lemons. Interestingly, many of the synthetic images seem to be more reminiscent of potatoes than lemons; given the nature of GANs, this is due to the model's generalisation of the dataset which contained lemons photographed at different angles - and so this generalisation of a shape is reflected in the produced images. A more uniform colour can be observed in healthy synthetic lemons, whereas the unhealthy lemons are given mould and dark styles, as well as several instances of gangrene. Similarly to the potato-like shape of the outputs, this could be attributed to the generalisation nature of GANs, several patterns have been observed during training, and these patterns are then applied while generating new images. 
\begin{figure}
    \centering
    \includegraphics[scale=0.8]{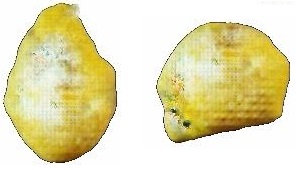}
    \caption{Two synthetic images of lemons belonging to the ``unhealthy" class generated by the Conditional GAN at full resolution. The lemons are seemingly presenting with symptoms of both mould and gangrene, showing that the generator model has learnt to generate fruit with undesirable characteristics. An unrealistic 'checker-boarding' effect has also been generated by the Conditional GAN on the surface of the fruit.
    }
    \label{fig:mouldy_synthetic_lemon}
\end{figure}
A higher resolution example of synthetic lemon images showing features of mould and gangrene can be observed in Figure \ref{fig:mouldy_synthetic_lemon}. The generator has seemingly learnt to cast light and shadows on the fruit as well as undesirable characteristics; both fruits have a texture and colour similar to mould on the surface, and the second fruit seems to have a gangrenous darker patch towards the top. 

\begin{figure}
    \centering
    \includegraphics[scale=0.75]{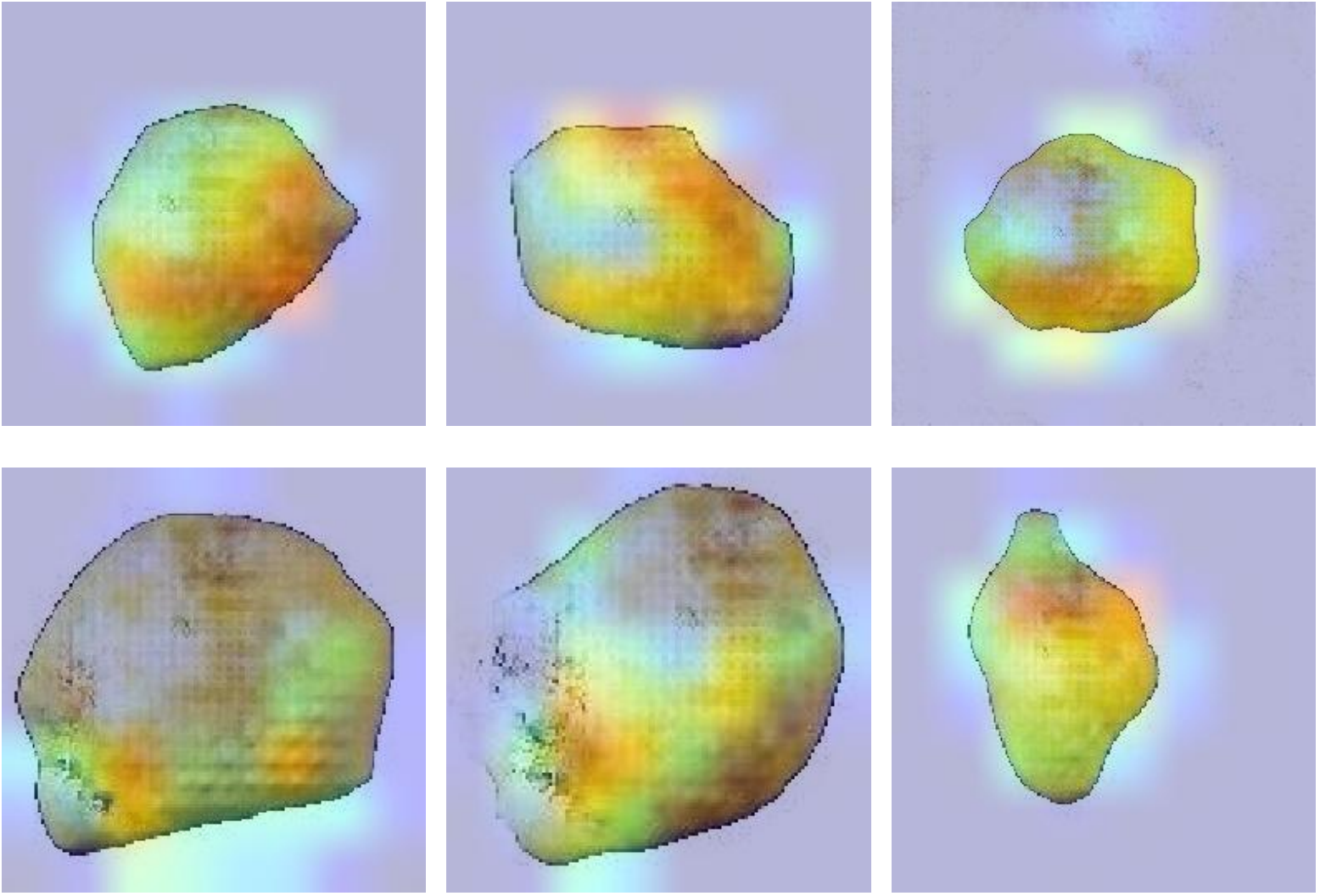}
    \caption{Grad-CAM analysis of six outputs from the Conditional GAN. Top row shows images belonging to the ``healthy" class and bottom row shows images belonging to the ``unhealthy" class. This Grad-CAM VGG16 CNN is trained only on real photographs, and no training has been performed on synthetic images.}
    \label{fig:gradcam_analysis}
\end{figure}
Figure \ref{fig:gradcam_analysis} shows class activation maps on six synthetic images (three per class) from the convolutional neural network trained \textbf{only} on the real data. All six images were predicted to belong to their ground-truth classes. Note that on the bottom row the Convolutional Neural Network focuses on issues on the flesh of the fruit such as mould in the first two images and a dark patch which may indicate gangrene on the right-most bottom image. Additionally, the class activation maps for the healthy fruit exist more generally around the shape. These behaviours are seemingly reminiscent of how a human would analyse the images, either focusing on unhealthy characteristics if the fruit is bad or observing the general image when the fruit shows no undesirable features. 

This analysis further shows that both desirable and undesirable characteristics are generalised, and, to an extent reproduced by the generative model. Indeed, the synthetic images are not perfect (as can be observed when comparing them via Figure \ref{fig:real_gan_examples}), but these activation maps provide insight into the useful synthetic knowledge existing within them.

\subsection{Classification Comparison}
\begin{figure}
    \centering
    \includegraphics[scale=0.7]{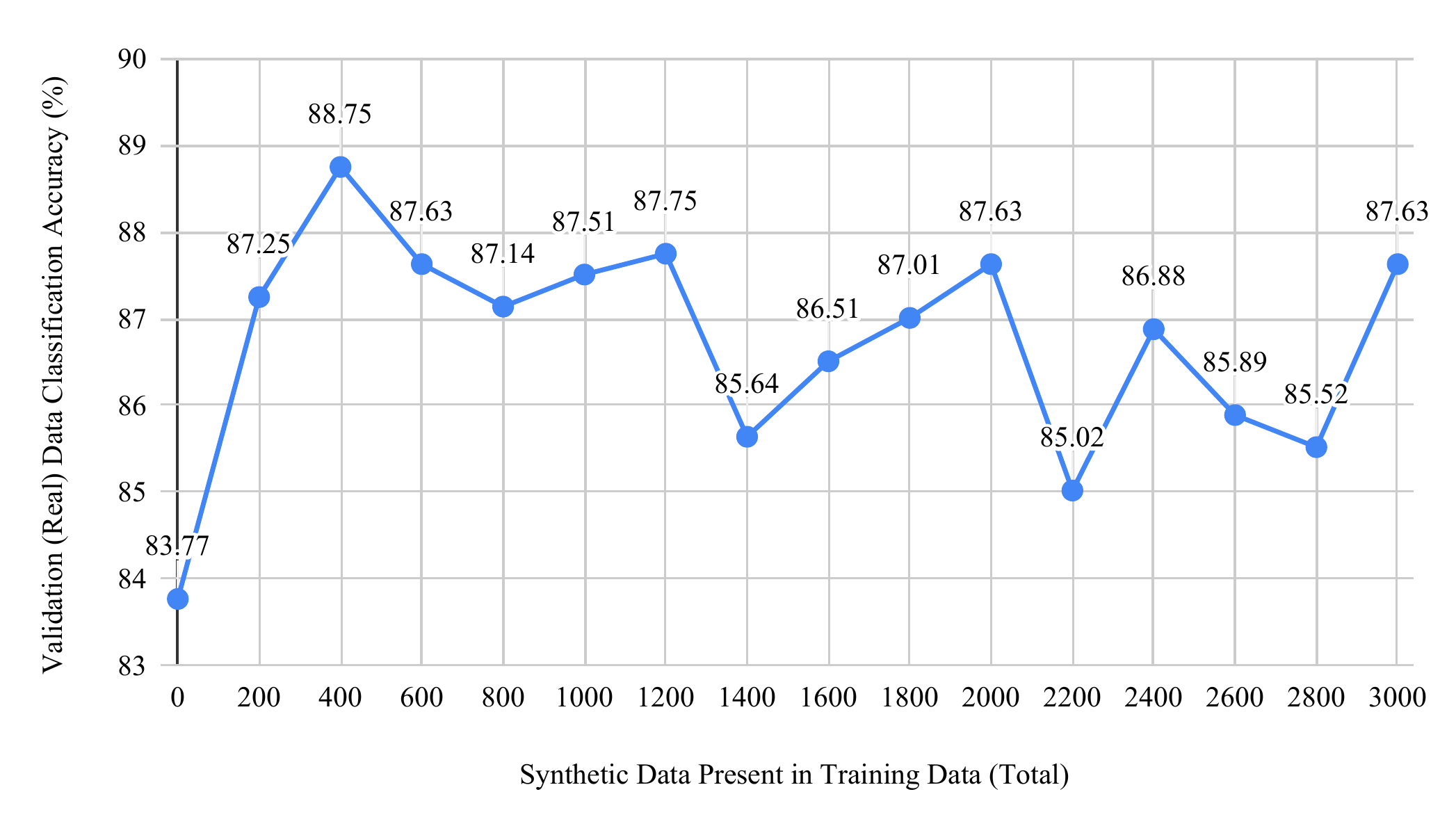}
    \caption{Comparison of how the vanilla CNN (no augmentation) performs against the models which have additional synthetic training data present.}
    \label{fig:gan_classification_results}
\end{figure}

\begin{table}[]
\centering
\caption{Comparison of non-augmented (first row) to training data augmentation with Conditional GAN-generated images.
}
\label{tab:final-results}
\footnotesize
\begin{tabular}{@{}rrr@{}}
\toprule
\multicolumn{1}{l}{\textbf{Synthetic Images Augmented (per Class)}} & \multicolumn{1}{l}{\textbf{Synthetic Data (Total)}} & \multicolumn{1}{l}{\textbf{Classification Accuracy (\%)}} \\ \midrule
0                                                                   & 0                                                   & 83.77                                                     \\
100                                                                 & 200                                                 & 87.25                                                     \\
200                                                                 & 400                                                 & \textbf{88.75}                                            \\
300                                                                 & 600                                                 & 87.63                                                     \\
400                                                                 & 800                                                 & 87.14                                                     \\
500                                                                 & 1000                                                & 87.51                                                     \\
600                                                                 & 1200                                                & 87.75                                                     \\
700                                                                 & 1400                                                & 85.64                                                     \\
800                                                                 & 1600                                                & 86.51                                                     \\
900                                                                 & 1800                                                & 87.01                                                     \\
1000                                                                & 2000                                                & 87.63                                                     \\
1100                                                                & 2200                                                & 85.02                                                     \\
1200                                                                & 2400                                                & 86.88                                                     \\
1300                                                                & 2600                                                & 85.89                                                     \\
1400                                                                & 2800                                                & 85.52                                                     \\
1500                                                                & 3000                                                & 87.63                                                     \\ \bottomrule
\end{tabular}
\end{table}

The results from the best CNN experiment along with the augmentation approaches can be observed in Figure \ref{fig:gan_classification_results} and Table \ref{tab:final-results}. All augmentation approaches scored higher than training only on the real images, showing that augmentation has had a positive effect on the learning process.The best set of results were achieved augmenting the dataset with 400 images (13.51\% of the whole dataset, 200 per class), which scored a classification accuracy of 88.75\% when classifying unseen images of both healthy and unhealthy lemons.  

Although there was a varying number of classification accuracies recorded, note that even the weakest augmentation approach (2200 images) caused the image recognition ability to rise from 83.77\% to 85.02\%. That is, of the 15 trials performed, all augmentation approaches outperformed the vanilla CNN. These results thus argue that Conditional GAN-based training data augmentation is a promising approach to improve fruit quality image classification. The generator model and weights are made publicly available for further exploration. 

\subsection{Pruning}
%
\begin{table}[]
\centering
\caption{Comparison of models when pruning with polynomial decay (for smaller model sizes) for both the vanilla and data augmentation approaches.}
\label{tab:prune-table}
\footnotesize
\begin{tabular}{@{}llll@{}}
\toprule
\multirow{2}{*}{\textbf{Pruned model size (\% of original)}} & \multirow{2}{*}{\textbf{Final Sparsity}} & \multicolumn{2}{l}{\textbf{Post-pruning Classification Accuracy (\%)}} \\ \cmidrule(l){3-4} 
                                                             &                                          & \textit{\textbf{Vanilla}}         & \textit{\textbf{Augmented (200 synthetic images)}}        \\ \cmidrule(r){1-2}
10                                                           & 0.9                                      & 60.1                              & 60.1                               \\
20                                                           & 0.8                                      & 73.85                             & 73.98                              \\
30                                                           & 0.7                                      & 75.71                             & 79.18                              \\
40                                                           & 0.6                                      & 77.45                             & 78.31                              \\
50                                                           & 0.5                                      & 80.79                             & 81.16                              \\
60                                                           & 0.4                                      & 81.54                             & 82.16                              \\
70                                                           & 0.3                                      & 82.16                             & 82.65                              \\
80                                                           & 0.2                                      & 83.64                             & 84.76                              \\
90                                                           & 0.1                                      & 82.9                              & 83.51                              \\
100                                                          & 0                                        & 83.77                             & 88.75                              \\ \bottomrule
\end{tabular}
\end{table}
\begin{figure}
    \centering
    \includegraphics[scale=0.7]{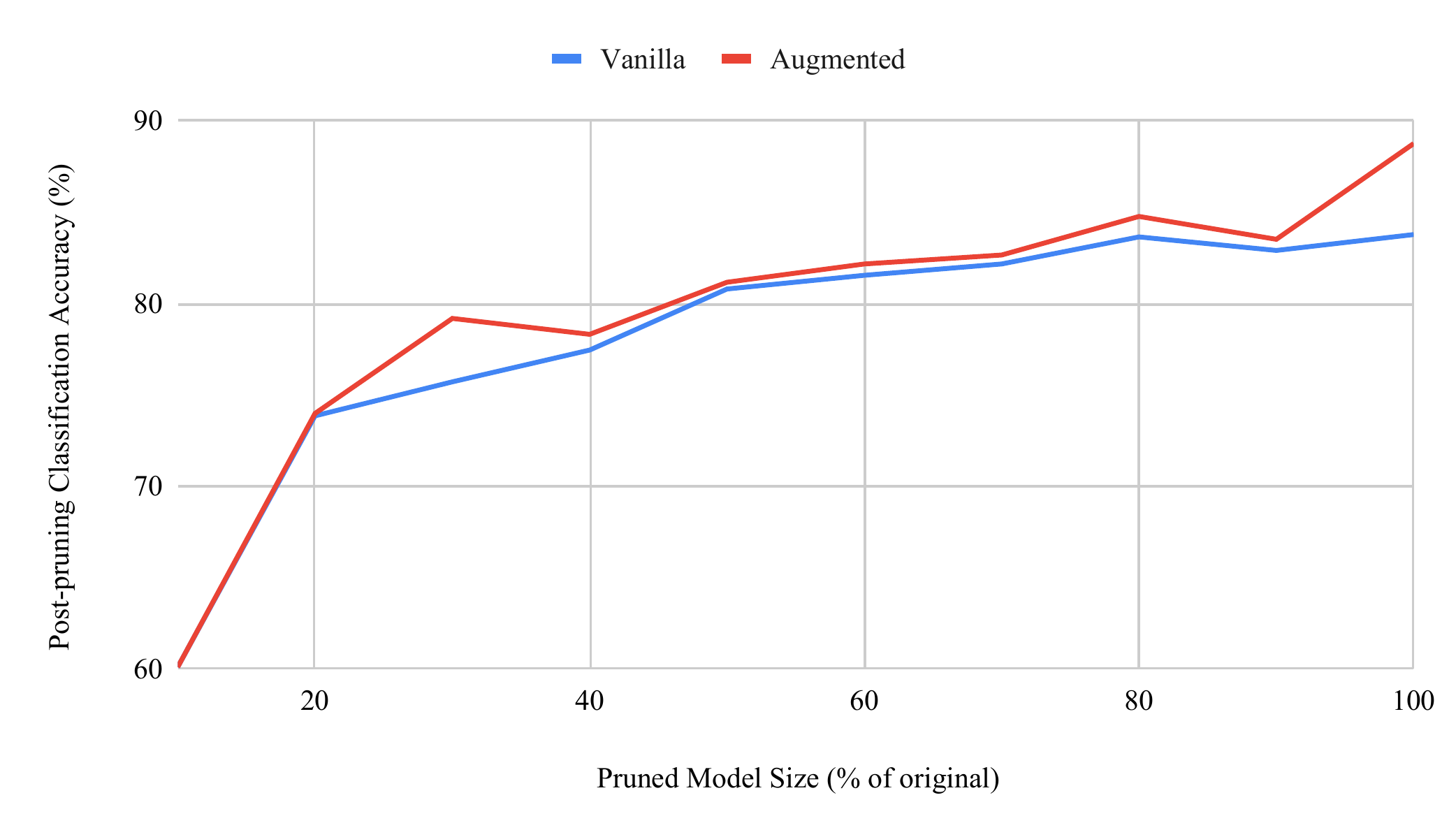}
    \caption{Comparison of the polynomial decay pruning accuracies for both vanilla and data augmentation models.}
    \label{fig:pruning-graph}
\end{figure}

Table \ref{tab:prune-table} and Figure \ref{fig:pruning-graph} show a comparison of the pruning experiments from 90\% to 10\% compression (through sparsity \textit{i.e.} zeroed weights). Although the final two models with all neurons present the best results, we note earlier that a faster model would be ideal for the problem at hand, since fruits are processed in large quantities and the models might need being ran under time and energy restrictions. It is interesting to note from these results that the data augmented network tends to outpeform the vanilla network in all cases. It is also interesting to note that a data augmentation network compressed to half of its original size can still perform at an accuracy of 81.16\%, effectively doubling production at a loss of 7.59\% accuracy.

\section{Conclusion and Future Work}
To finally conclude, we first noted that autonomous sorting of healthy fruit from undesirable fruit is possible through contemporary computer vision technologies. We explored the concepts of fine-tuning, transfer learning, and Conditional GAN-based training data augmentation; our results showed that the recognition (and thus sorting) of fruit images was improved when augmentation was introduced. We found that introducing 400 synthetic data points had the largest impact, raising the recognition accuracy from 83.77\% to 88.75\% via a convolutional neural network. Finally, we performed Grad-CAM analysis with the model trained only on real photographs to show that the Conditional GAN was successful in imagining the undesirable characteristics on the flesh of the synthetic fruit generated. Thus, this work argues that Conditional Generative Adversarial Networks have the ability to produce new data to alleviate issues of data scarcity in the problem of fruit health classification.

Although we explored the lowest loss of the generator compared to the final loss, and a vast improvement was made albeit with the said higher loss, future further exploration of model weights throughout the training process may be useful to explore. Given the behaviour of losses spiking, several of the later models should be explored in order to ascertain whether the final epoch did produce the best model for this approach. Although the Conditional GAN-based approach showed promise to the classical CNN, one limitation of this work is that we compare the best train-test CNN to the Conditional GAN; in future, given resource availability to quickly train several Conditional GANs, a better metric for evaluation would be to train $k$ Conditional GANs during $k$-fold cross validation. This would allow for better scientific accuracy. Note that the Conditional GAN was trained for 17 hours on a leading GPU, and thus such a solution would only likely be viable in the future with better technology. 

Pruning showed that models could be presented at a fraction of their original size and only lose a small amount of classification ability. With this in mind, other pruning techniques could be explored in future to further maximise the model's ability in terms of real-world usage. 

Some future work is also needed for the real-world use of this approach. The nature of the dataset that these models are trained with dictate that image segmentation is required in order to separate the fruit from the background (such as a surface or other fruit). If a segmentation network is trained and applied prior to the preprocessing described in these techniques, then the approach can be tested on more data in a much higher volume, i.e., fruit production. 

\section{Model and Code Availability}
For replicability purposes as well as future research, the generator model and synthetic image generation Python code have been made available at: \url{https://github.com/jordan-bird/synthetic-fruit-image-generator}.

\bibliographystyle{ieeetr}  
\bibliography{references}  

\end{document}